\pdfoutput=1
\documentclass[11pt]{article}

\usepackage[final]{acl}
\usepackage{times}
\usepackage{latexsym}

\usepackage[T1]{fontenc}
\usepackage[utf8]{inputenc}

\usepackage{microtype}

\usepackage{inconsolata}
\usepackage{amsmath}
\usepackage{lipsum}
\usepackage{booktabs}
\usepackage{float}
\usepackage{tabularx}
\usepackage{tikz}
\usepackage{pgfplots}
\usepgfplotslibrary{groupplots}
\usepackage{xfrac}
\usepackage{amssymb}
\usepackage{enumitem}

\usepackage{todonotes}
\usepackage{natbib}
\usepackage{mathtools}

\definecolor{uzhblue}{HTML}{0C279E}
\definecolor{uzhorange}{HTML}{E95A29}

\newcommand{\chrf}{\textsc{ChrF}}
\newcommand{\comet}{\textsc{COMET}}
\newcommand{\comettt}{\mbox{\comet{}-22}}
\newcommand{\cometinho}{\mbox{Cometinho}}
\newcommand{\bleurt}{\textsc{BLEURT}}
\newcommand{\bleurtt}{\mbox{\textsc{BLEURT-20}}}
\newcommand{\cometxl}{\mbox{\textsc{xCOMET-XL}}}

\DeclareMathOperator*{\optop}{top-T}
\DeclareMathOperator*{\argmax}{arg\,max} %
\newcommand{\bigO}{O}

\title{Linear-time Minimum Bayes Risk Decoding with Reference Aggregation}

\author{Jannis Vamvas \and Rico Sennrich\\
  Department of Computational Linguistics, University of Zurich\\
  \texttt{\{vamvas,sennrich\}@cl.uzh.ch}}

\begin{document}
\maketitle
\begin{abstract}
Minimum Bayes Risk (MBR) decoding is a text generation technique that has been shown to improve the quality of machine translations, but is expensive, even if a sampling-based approximation is used.
Besides requiring a large number of sampled sequences, it requires the pairwise calculation of a utility metric, which has quadratic complexity.
In this paper, we propose to approximate pairwise metric scores with scores calculated against aggregated reference representations.
This changes the complexity of utility estimation from $\bigO(n^2)$ to $\bigO(n)$, while empirically preserving most of the quality gains of MBR decoding.
We release our source code.\footnote{\url{https://github.com/ZurichNLP/mbr}}
\end{abstract}

\section{Introduction}
The idea of generating translations by maximizing a metric of translation quality~\cite{kumar-byrne-2004-minimum} has recently been revived in the context of neural machine translation.
In sampling-based MBR decoding~\cite{eikema-aziz-2020-map}, many hypotheses are sampled from the model distribution, and their expected utility is estimated using Monte Carlo~(MC) sampling.
This approach has been shown to improve translation quality compared to beam search, especially when neural metrics are used for utility estimation~\cite{freitag-etal-2022-high}.

Estimating utility through MC sampling has quadratic complexity in the number of samples, which limits practical application.
Previous work suggested pruning the number of samples based on a cheaper metric or a smaller number of references~\cite{eikema-aziz-2022-sampling,cheng-vlachos-2023-faster}.
In this paper, we propose \textit{reference aggregation}, an alternative efficiency technique that exploits the fact that most common metrics represent text sequences in averageable form, e.g., as n-gram statistics or as embeddings.
Specifically, we combine representations of the references into an aggregate reference representation, which we then use for utility estimation.
Our proposed approximation still relies on MC sampling, but on a lower level:
Rather than computing an MC estimate of the expected utility, we compute an MC estimate of the ``true'' reference representation in the feature space of the given utility metric.
Since this estimate only needs to be computed once, our approach has linear complexity in the number of sampled hypotheses and references.

We report empirical results for four translation directions and two utility metrics: \chrf{}~\cite{popovic-2015-chrf}, which is based on character n-gram overlap, and \comet{}~\cite{rei-etal-2020-comet}, a neural network trained with examples of human translation quality judgments.
For \chrf{}, we find that reference aggregation reduces the time needed for computing the utility of 1024~samples by 82.0\%, without affecting translation quality.
For \comet{}, metric accuracy does decrease with aggregation, but to a lesser extent than with simply reducing the number of references.
Depending on the \comet{} model, computation time is reduced by 95–99\%, which makes reference aggregation an efficient method for hypothesis pruning with \comet{}.

\section{Background and Related Work}
Sampling-based MBR~\cite{eikema-aziz-2020-map} selects a translation $\textit{hyp}^*$ out of a set of translation hypotheses $\textit{hyp}_1, \ldots, \textit{hyp}_n \in \textit{hyps}$ by maximizing~(expected) utility:
\begin{equation}
    \textit{hyp}^* = \argmax_{\textit{hyp} \, \in \, \textit{hyps}} \text{utility}(\textit{hyp}).
\end{equation}

The set of hypotheses is sampled from the model distribution $p(\textit{hyp}|\textit{src})$.
\citet{eikema-aziz-2020-map} propose to approximate the utility using MC sampling: sample a set of pseudo-references $\textit{refs} = \{\textit{ref}_1, \ldots, \textit{ref}_m\} \sim p(\textit{ref} |\textit{src})$ from the model and calculate a metric against each sampled reference:
\begin{equation}
    \text{utility}(\textit{hyp}) \approx \frac{1}{m} \sum_{\textit{ref} \, \in \, \textit{refs}} \text{metric}(\textit{hyp}, \textit{ref}).
\end{equation}

For machine translation, typical such metrics are \chrf{}~\cite{popovic-2015-chrf} and BLEU~\cite{papineni-etal-2002-bleu}, which are based on n-gram statistics, or neural metrics such as \comet{}~\cite{rei-etal-2020-comet} and \bleurt{}~\cite{sellam-etal-2020-bleurt}.

A line of research has focused on improving the efficiency of sampling-based MBR.
\citet{eikema-aziz-2022-sampling} propose \textit{coarse-to-fine MBR}, which prunes the hypotheses based on a cheaper metric, and \textit{N-by-S MBR}, which uses fewer references than hypotheses.
\citet{cheng-vlachos-2023-faster} propose \textit{confidence-based pruning}, where the number of hypotheses is iteratively reduced based on an increasing number of references.
\citet{jinnai2024hyperparameterfree} interpret sampling-based MBR as an instance of \textit{medoid identification} and apply an established approximation algorithm to this problem.
A line of work uses MBR outputs as a training reward, avoiding the inefficiency of MBR during deployment~\cite{finkelstein2023mbr,yang2023direct}.
Finally, alternative reranking approaches that do not require pairwise comparisons have been proposed~\cite{fernandes-etal-2022-quality}.

Several other works investigate the aggregation of reference representations to develop a faster variant of MBR decoding.
\citet{denero-etal-2009-fast} perform reference aggregation in the context of statistical machine translation~(SMT). Since SMT does not afford random sampling of pseudo-references, they aggregate references from translation forests or $k$-best lists. Our study shows the effectiveness of reference aggregation from sampled pseudo-references, and for neural metrics such as COMET.
Furthermore, concurrent to our work, \citet{deguchi-etal-2024-centroid} propose to aggregate the sentence embeddings of COMET, and use $k$-means to group the references into multiple clusters.
%While their work focuses on finding an optimal clustering of COMET embeddings for MBR decoding, this paper focuses on an evaluation of straightforward reference aggregation with COMET and ChrF, which in the latter case is a lossless procedure.

\section{Reference Aggregation}
Our approach is based on the observation that most metrics that are commonly used for MBR make use of feature representations that can be aggregated.
For example, the n-gram statistics used by \chrf{} can be aggregated by averaging the counts of the n-grams across all references; and the sentence embeddings used by \comet{} can be aggregated by calculating an average sentence embedding.

For simplicity, we re-use the above notation, where $\textit{hyp}$ is a hypothesis and $\textit{ref}$ is a reference, but we now assume that they are represented in an averageable form.
We then combine the set of references $\textit{refs}$ into an aggregate representation $\overline{\textit{ref}}$:
\begin{equation}
    \overline{\textit{ref}} = \frac{1}{m} \sum_{\textit{ref} \, \in \, \textit{refs}} \textit{ref}.
\end{equation}
We approximate the expected utility of a sampled hypothesis by calculating a single metric score against this aggregate representation:
\begin{equation}
    \text{utility}(\textit{hyp}) \approx \text{metric}(\textit{hyp}, \overline{\textit{ref}}).
\end{equation}
Like with standard sampling-based MBR, it is possible to interpret this approximation as MC sampling:
By averaging over representations of sampled references, we estimate a representation of the ``true'' reference, which we then use for approximating the expected utility of each sampled hypothesis.
Importantly, the computational complexity of our approach is in $\bigO(|\textit{hyps}|+|\textit{refs}|)$ rather than $\bigO(|\textit{hyps}| \cdot |\textit{refs}|)$; see Appendix~\ref{sec:appendix-complexity} for a discussion.

\subsection{Application to chrF Metric}

\chrf{}~\cite{popovic-2015-chrf} is defined as an F-score over character n-grams:
\begin{equation}
    \textsc{ChrF}_\beta = \frac{(1 + \beta^2) \cdot \textsc{ChrP} \cdot \textsc{ChrR}}{\beta^2 \cdot \textsc{ChrP} + \textsc{ChrR}},
\end{equation}
where
\[
\textsc{ChrP} = \frac{|\textit{hyp} \cap \textit{ref}|}{|\textit{hyp}|} ~~ \text{and} ~~ \textsc{ChrR} = \frac{|\textit{hyp} \cap \textit{ref}|}{|\textit{ref}|},
\]
and the parameter \(\beta\) controls the relative importance of precision and recall.
The representations \/\textit{hyp} and \textit{ref} are bags of n-grams, i.e., objects that map each n-gram to its count in the string.

We apply reference aggregation to \chrf{} by averaging the counts of n-grams across all references:
\begin{equation}\label{eq:pairwise-chrf}
    \overline{\textit{ref}} = \frac{1}{m} \biguplus_{\textit{ref} \, \in \, \textit{refs}} \textit{ref},
\end{equation}
where \(\biguplus\) is an operation that sums up the counts of each n-gram.
We then approximate the expected utility of a hypothesis by calculating $\textsc{ChrF}_\beta(\textit{hyp}, \overline{\textit{ref}})$.
Appendix~\ref{appendix:chrf} provides a more formal definition of reference aggregation for \chrf{}.

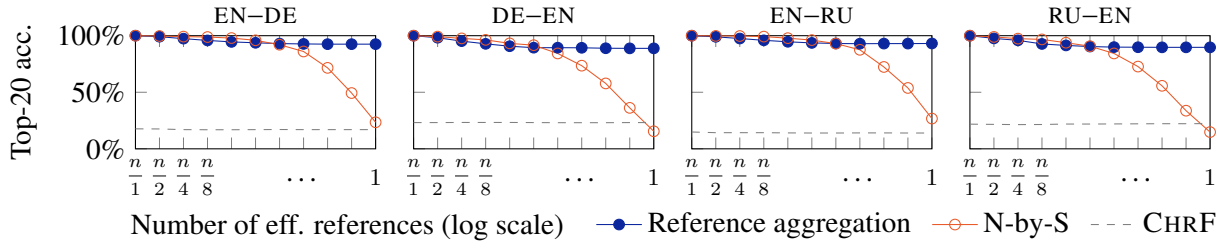
\begin{figure*}[htb!]
\hspace{0.1cm}
\textbf{Accuracy of efficiency methods with \chrf{} as utility metric}
\vspace{-0.1cm}
\begin{tikzpicture}
\pgfplotsset{width=0.1975\textwidth}
\begin{groupplot}[
    group style={
        group size=4 by 1,
        xlabels at=edge bottom,
        ylabels at=edge left,
        horizontal sep=0.5cm,
    },
    scale only axis,
    xmode=log,
    log basis x={2},
    xlabel={Number of effective references (log scale)},
    xlabel style={xshift=1.6cm, yshift=-0.2cm},
    xmin=1, xmax=1024,
    xtick={1,2,4,8,16,32,64,128,256,512,1024},
    xticklabels={\scriptsize{$\dfrac{n}{1}$}, \scriptsize{$\dfrac{n}{2}$}, \scriptsize{$\dfrac{n}{4}$}, \scriptsize{$\dfrac{n}{8}$},,,,\vphantom{\scriptsize{$\dfrac{n}{}$}}\ldots,,,\vphantom{\small{$\dfrac{n}{n}$}}\small{$1$}},
    ymin=0, ymax=1,
    ytick={0, 0.5, 1},
    yticklabels={0\%, 50\%, 100\%},
    grid style=dashed,
    height=1.5cm
]

% Plot 1 (en-de accuracy)
\nextgroupplot[
    title=\textsc{en--de},
    title style={yshift=-5pt},
    ylabel={Top-20 acc.},
    ylabel near ticks,
    ylabel style={align=center},
    ytick pos=left,
    legend style={at={(3.52,-0.874)},anchor=south, draw=none, fill=none, style={/tikz/every even column/.append style={column sep=0.2cm}}},
    legend columns=-1
]
\addplot[color=uzhblue, mark=*] coordinates {
(1,1.00000)(2,0.99800)(4,0.99301)(8,0.99202)(16,0.99102)(32,0.99202)(64,0.99202)(128,0.99202)(256,0.99202)(512,0.99202)(1024,0.99202)
};
\addlegendentry{Reference aggregation}

\addplot[color=uzhorange, mark=o] coordinates {
(1,1.00000)(2,0.98703)(4,0.98303)(8,0.95509)(16,0.93014)(32,0.87824)(64,0.74850)(128,0.61277)(256,0.39421)(512,0.20659)(1024,0.10180)
};
\addlegendentry{N-by-S}

% Plot 2 (de-en accuracy)
\nextgroupplot[title=\textsc{de--en}, title style={yshift=-5pt}, yticklabels={}, ytick pos=left, xlabel=\empty]
\addplot[color=uzhblue, mark=*] coordinates {
(1,1.00000)(2,0.98900)(4,0.98600)(8,0.98400)(16,0.98400)(32,0.98200)(64,0.98100)(128,0.98100)(256,0.98100)(512,0.98200)(1024,0.98200)
};
\addplot[color=uzhorange, mark=o] coordinates {
(1,1.00000)(2,0.98000)(4,0.95400)(8,0.91900)(16,0.87600)(32,0.79100)(64,0.69300)(128,0.50400)(256,0.34800)(512,0.20400)(1024,0.12500)
};

% Plot 3 (en-ru accuracy)
\nextgroupplot[title=\textsc{en--ru}, title style={yshift=-5pt}, yticklabels={}, ytick pos=left, xlabel=\empty]
\addplot[color=uzhblue, mark=*] coordinates {
(1,1.00000)(2,0.99601)(4,0.99301)(8,0.99102)(16,0.99102)(32,0.99102)(64,0.98902)(128,0.98902)(256,0.98902)(512,0.98902)(1024,0.98902)
};
\addplot[color=uzhorange, mark=o] coordinates {
(1,1.00000)(2,0.99601)(4,0.98403)(8,0.96108)(16,0.93014)(32,0.86028)(64,0.74750)(128,0.57685)(256,0.34930)(512,0.23154)(1024,0.12575)
};

% Plot 4 (ru-en accuracy)
\nextgroupplot[title=\textsc{ru--en}, title style={yshift=-5pt}, yticklabels={}, ytick pos=left, xlabel=\empty]
\addplot[color=uzhblue, mark=*] coordinates {
(1,1.00000)(2,0.99100)(4,0.98400)(8,0.98300)(16,0.98000)(32,0.97800)(64,0.97800)(128,0.97700)(256,0.97500)(512,0.97500)(1024,0.97400)
};
\addplot[color=uzhorange, mark=o] coordinates {
(1,1.00000)(2,0.97200)(4,0.95800)(8,0.92800)(16,0.87200)(32,0.80600)(64,0.67300)(128,0.53900)(256,0.35800)(512,0.23300)(1024,0.11500)
};
\end{groupplot}
\end{tikzpicture}
\vspace{-0.9cm}
\hspace{0.1cm}
\textbf{Accuracy of efficiency methods with \comettt{} as utility metric}
\vspace{-0.1cm}
\begin{tikzpicture}
\pgfplotsset{width=0.1975\textwidth}
\begin{groupplot}[
    group style={
        group size=4 by 1,
        xlabels at=edge bottom,
        ylabels at=edge left,
        horizontal sep=0.5cm,
    },
    scale only axis,
    xmode=log,
    log basis x={2},
    xlabel={Number of eff. references (log scale)},
    xlabel style={xshift=1.21cm, yshift=-0.2cm},
    xmin=1, xmax=1024,
    xtick={1,2,4,8,16,32,64,128,256,512,1024},
    xticklabels={\scriptsize{$\dfrac{n}{1}$}, \scriptsize{$\dfrac{n}{2}$}, \scriptsize{$\dfrac{n}{4}$}, \scriptsize{$\dfrac{n}{8}$},,,,\vphantom{\scriptsize{$\dfrac{n}{}$}}\ldots,,,\vphantom{\small{$\dfrac{n}{n}$}}\small{$1$}},
    ymin=0, ymax=1,
    ytick={0, 0.5, 1},
    yticklabels={0\%, 50\%, 100\%},
    grid style=dashed,
    height=1.5cm
]

% Plot 1 (en-de accuracy)
\nextgroupplot[
    title=\textsc{en--de},
    title style={yshift=-5pt},
    ylabel={Top-20 acc.},
    ylabel near ticks,
    ylabel style={align=center},
    ytick pos=left,
    legend style={at={(3.22,-0.874)},anchor=south, draw=none, fill=none, style={/tikz/every even column/.append style={column sep=0.2cm}}},
    legend columns=-1
]
\addplot[color=uzhblue, mark=*] coordinates {
(1,1.00000)(2,0.99102)(4,0.97405)(8,0.95808)(16,0.94511)(32,0.93713)(64,0.93014)(128,0.92715)(256,0.92515)(512,0.92515)(1024,0.92515)
};
\addlegendentry{Reference aggregation}

\addplot[color=uzhorange, mark=o] coordinates {
(1,1.00000)(2,0.99701)(4,0.99202)(8,0.98802)(16,0.97904)(32,0.95908)(64,0.92016)(128,0.86028)(256,0.71557)(512,0.49301)(1024,0.23453)
};
\addlegendentry{N-by-S}

\addplot[dashed, gray] coordinates {
(1,0.17665)(2,0.17565)(4,0.16966)(8,0.16866)(16,0.16866)(32,0.16966)(64,0.17066)(128,0.16966)(256,0.16966)(512,0.16966)(1024,0.16966)
};
\addlegendentry{\chrf{}}

% Plot 2 (de-en accuracy)
\nextgroupplot[title=\textsc{de--en}, title style={yshift=-5pt}, yticklabels={}, ytick pos=left, xlabel=\empty]
\addplot[color=uzhblue, mark=*] coordinates {
(1,1.00000)(2,0.98200)(4,0.95200)(8,0.92800)(16,0.90700)(32,0.89400)(64,0.89400)(128,0.89300)(256,0.88900)(512,0.88800)(1024,0.88800)
};
\addplot[color=uzhorange, mark=o] coordinates {
(1,1.00000)(2,0.99200)(4,0.97600)(8,0.96300)(16,0.93200)(32,0.91600)(64,0.84300)(128,0.73500)(256,0.57800)(512,0.36200)(1024,0.15500)
};
\addplot[dashed, gray] coordinates {
(1,0.23200)(2,0.23200)(4,0.23300)(8,0.23300)(16,0.23300)(32,0.23200)(64,0.23100)(128,0.23100)(256,0.23100)(512,0.23200)(1024,0.23200)
};

% Plot 3 (en-ru accuracy)
\nextgroupplot[title=\textsc{en--ru}, title style={yshift=-5pt}, yticklabels={}, ytick pos=left, xlabel=\empty]
\addplot[color=uzhblue, mark=*] coordinates {
(1,1.00000)(2,0.99002)(4,0.97505)(8,0.95908)(16,0.94611)(32,0.93613)(64,0.93114)(128,0.92914)(256,0.93014)(512,0.93014)(1024,0.93114)
};
\addplot[color=uzhorange, mark=o] coordinates {
(1,1.00000)(2,0.99800)(4,0.99701)(8,0.99202)(16,0.97804)(32,0.96108)(64,0.93014)(128,0.87525)(256,0.72355)(512,0.53792)(1024,0.26747)
};
\addplot[dashed, gray] coordinates {
(1,0.14870)(2,0.14271)(4,0.14271)(8,0.14271)(16,0.14072)(32,0.13972)(64,0.13972)(128,0.14072)(256,0.14072)(512,0.13972)(1024,0.13972)
};

% Plot 4 (ru-en accuracy)
\nextgroupplot[title=\textsc{ru--en}, title style={yshift=-5pt}, yticklabels={}, ytick pos=left, xlabel=\empty]
\addplot[color=uzhblue, mark=*] coordinates {
(1,1.00000)(2,0.97600)(4,0.95900)(8,0.92700)(16,0.91400)(32,0.90400)(64,0.89900)(128,0.89800)(256,0.89700)(512,0.89700)(1024,0.89700)
};
\addplot[color=uzhorange, mark=o] coordinates {
(1,1.00000)(2,0.98900)(4,0.97400)(8,0.96700)(16,0.93800)(32,0.90700)(64,0.84300)(128,0.72700)(256,0.55700)(512,0.33700)(1024,0.14800)
};
\addplot[dashed, gray] coordinates {
(1,0.21800)(2,0.21600)(4,0.21400)(8,0.21500)(16,0.21900)(32,0.21900)(64,0.22000)(128,0.22100)(256,0.22200)(512,0.22200)(1024,0.22300)
};

\end{groupplot}
\end{tikzpicture}
\vspace{-1.1cm}
\caption{
How accurately do MBR efficiency methods approximate standard MBR?
In this validation experiment on \textit{newstest21}, we gradually increase efficiency by using fewer references for pairwise utility estimation – either by subsampling the references (N-by-S; \citealp{eikema-aziz-2022-sampling}) or by aggregating their representations using partial aggregation~(Section~\ref{sec:partial-aggregation}).
We report top-20 accuracy, which describes how often an efficiency method ranks the correct hypothesis (as selected by standard MBR) among the top 20 hypotheses.
An efficiency method with a high top-20 accuracy could be used for pruning the number of hypotheses to 20 before standard MBR is applied.
}
\label{fig:validation-top20-accuracy}
\end{figure*}

\subsection{Application to COMET Metric}

\comet{}~\cite{rei-etal-2020-comet} is a pre-trained Transformer model~\cite{vaswani2017attention} that has been fine-tuned to predict human judgments of translation quality.
In this paper, we focus on the Estimator model architecture, which directly estimates a quality score given a hypothesis, a reference and the source sequence.
\comet{} separately encodes these three inputs into fixed-size embeddings:
\begin{align*}
\textbf{\textit{hyp}},~\textbf{\textit{ref}},~\textbf{\textit{src}} &= \text{emb}(\textit{hyp}),~\text{emb}(\textit{ref}),~\text{emb}(\textit{src}).
\end{align*}
The three embeddings are then fed into a feed-forward module, which outputs a scalar score:
\begin{equation}\label{eq:comet-score}
    \text{comet}(\textit{hyp}) = \text{score}(\textbf{\textit{hyp}},~\textbf{\textit{ref}},~\textbf{\textit{src}}).
\end{equation}
We apply reference aggregation to \comet{} by averaging the reference embeddings:
\begin{equation}
    \overline{\textbf{\textit{ref}}} = \frac{1}{m} \sum_{\textit{ref} \, \in \, \textit{refs}} \text{emb}(\textit{ref}),
\end{equation}
calculating a single score per hypothesis:
\begin{equation}
    \text{comet}(\textit{hyp}) \approx \text{score}(\textbf{\textit{hyp}},~\overline{\textbf{\textit{ref}}},~\textbf{\textit{src}}).
\end{equation}

\subsection{Partial Aggregation}
\label{sec:partial-aggregation}
To better understand the loss of accuracy incurred by aggregation, we experiment with partial aggregation, where we vary the number of references that are combined into an average.
Given $m$ references and a desired number of references ~$s$ that should effectively be used for pairwise utility estimation, we partition the set of references into~$s$ subsets and create an aggregate reference for each subset.
Appendix~\ref{appendix:partial-aggregation} presents a formal description of partial aggregation.

\subsection{Aggregate-to-fine MBR}
\label{sec:aggregate-to-fine}
Analogously to \textit{coarse-to-fine MBR}~\cite{eikema-aziz-2022-sampling}, we evaluate an \textit{aggregate-to-fine MBR} approach.
Specifically, we use the aggregate reference to prune the number of hypotheses to~20 in a first step.
In a second step, we use standard MBR to select the best hypothesis from the pruned set.
A formal description is provided in Appendix~\ref{appendix:aggregate-to-fine}.

\begin{table*}[htb]
  \centering
\begin{tabularx}{\textwidth}{@{}Xrrrrr@{\hskip 0.3in}r@{}}
\toprule
& \textsc{en--de} & \textsc{de--en} & \textsc{en--ru} & \textsc{ru--en} & Avg. & Time (utility / total) \\
\midrule
Beam search (size 5) & 76.16 & 72.56 & 68.50 & 75.47 & 73.17 &  - ~/ \phantom{0}0.2\,s \\
Epsilon sampling ($\epsilon=0.02$) & 73.39 & 69.70 & 65.79 & 72.13 & 70.25 & - ~/ \phantom{0}0.2\,s \\
\midrule
\mbox{MBR with \chrf{} metric} &  &  &  &  &  &  \\
– standard MBR & 76.03 & 72.73 & 69.52 & 75.51 & 73.44 & \phantom{0}0.4\,s ~/ \phantom{0}5.2\,s \\
– reference aggregation & 75.95 & \textbf{72.79} & \underline{69.46} & \underline{75.45} & \underline{73.41} & \phantom{0}0.1\,s ~/ \phantom{0}4.9\,s \\
– aggregate-to-fine MBR & \underline{76.02} & \textbf{72.80} & \underline{69.54} & \underline{75.47} & \underline{73.46} & \phantom{0}0.2\,s ~/ \phantom{0}5.0\,s \\
\midrule
\mbox{MBR with \comettt{} metric} & & & & &  \\
– standard MBR & 77.64 & 73.57 & 72.40 & 76.11 & 74.93 & 23.1\,s ~/ 27.9\,s \\
– reference aggregation & 77.21 & 73.36 & 72.05 & \underline{76.05} & 74.67 & \phantom{0}1.1\,s ~/ \phantom{0}5.9\,s \\
– aggregate-to-fine MBR & 77.54 & \underline{73.52} & \underline{72.29} & \underline{76.13} & 74.87 & \phantom{0}1.5\,s ~/ \phantom{0}6.3\,s \\
\bottomrule
\end{tabularx}
  \caption{Test results on \textit{newstest22}, using \bleurtt{} for automatic evaluation.
  We use 1024 samples/references for MBR. In the last column, we report the average time needed for translating a segment, measuring (a) the time needed for utility estimation only, and (b) the total, end-to-end time needed for translation.
  \underline{Underline}: no significant BLEURT difference to standard MBR; \textbf{bold}: significantly better than standard MBR (bootstrap test, $p<0.05$).
  }
  \label{tab:test-results}
\end{table*}

\section{Experimental Setup}

\paragraph{Data}
We use \textit{newstest21}~\cite{akhbardeh-etal-2021-findings} as validation data and \textit{newstest22}~\cite{kocmi-etal-2022-findings} as test data.

\paragraph{Generation Parameters}
As baselines, we evaluate beam search with a beam size of 5 and epsilon sampling~\cite{hewitt-etal-2022-truncation} with $\epsilon=0.02$.
For MBR, we generate 1024 samples per segment using epsilon sampling and re-use the same samples as references.
While this approach does not guarantee that the estimation of the expected utility is unbiased~\cite{eikema-aziz-2022-sampling}, it has empirically been found to work well~\cite{freitag-etal-2023-epsilon}.

\paragraph{Models}
We use open-source NMT models trained for the \textsc{en--de}, \textsc{de--en}, \textsc{en--ru} and \textsc{ru--en} translation directions~\cite{ng-etal-2019-facebook}.\footnote{
The models were trained with a label smoothing of $\epsilon=0.1$~\cite{szegedy2016rethinking}, which is a common choice in NMT.
Some previous studies of MBR trained custom models without label smoothing~(e.g., \citealp{eikema-aziz-2020-map}).
We argue that this is only necessary if unbiased utility estimates are sought through ancestral sampling, and should be less of a concern with epsilon sampling.
}
The authors provide an ensemble of four models per direction, but we restrict our experiments to one single model per direction.
We use the \textit{Fairseq} codebase~\cite{ott-etal-2019-fairseq} for model inference.

\paragraph{Metrics}
For estimating the utilities with \chrf{}, we use a custom implementation of \chrf{}\footnote{\url{https://github.com/jvamvas/fastChrF}} that is equivalent to SacreBLEU~\cite{post-2018-call} with default settings\footnote{chrF2|\#:1|case:mixed|eff:yes|nc:6|nw:0|space:no|v:2.0.0}.
As \comet{} model, we use \comettt{}~\cite{rei-etal-2022-comet}; because this model was not trained on annotations of \textit{newstest21} or \textit{newstest22}, a train–test overlap can be ruled out.
We estimate wall-clock time based on a part of the segments, using a system equipped with an NVIDIA GeForce RTX 3090 and an AMD EPYC 7742 64-core processor.

\section{Results}

\subsection{Validation results}
Figure~\ref{fig:validation-top20-accuracy} evaluates how accurately MBR efficiency methods approximate standard MBR.
We report top-20 accuracy, motivated by the idea of coarse-to-fine MBR: any method with perfect \mbox{top-20 accuracy} could be used for pruning the hypothesis set to 20 without affecting quality.
Results for top-1 accuracy are reported in Appendix~\ref{sec:appendix-top1-accuracy}.\footnote{Accuracy was proposed by \citet{cheng-vlachos-2023-faster} as an evaluation metric for MBR efficiency methods.}

For \chrf{}, we observe that reference aggregation is Pareto superior to N-by-S, maintaining near-perfect top-20 accuracy even if a single aggregate reference is used.
For \comet{}, reference aggregation causes some loss of accuracy, but outperforms N-by-S if the number of effective references is $\leq 16$, where efficiency is highest.
In addition, we find that reference aggregation approximates standard (pairwise) \comet{} much better than using \chrf{} as a coarse metric does, providing a clear motivation for aggregate-to-fine MBR as an alternative to coarse-to-fine MBR.

\subsection{Test results}
In Table~\ref{tab:test-results}, we report test results for \textit{newstest22}, focusing on a comparison between fast baseline algorithms (beam search and sampling) and MBR (with or without reference aggregation).
We perform an automatic evaluation using \bleurtt{}~\cite{sellam-etal-2020-bleurt}, chosen because it is unrelated to the utility metrics we use for MBR\@.
\chrf{} and \comet{} scores are reported in Appendix~\ref{sec:appendix-test-results}.

The results show that reference aggregation narrows the efficiency gap between MBR and beam search while preserving most of the quality gain of standard MBR.
Reference aggregation speeds up utility estimation by 82.0\% for \chrf{} and 95.1\% for \comettt{}.
For \comettt{}, this reduces the total time needed for translation by 78.8\%; for \chrf{}, pairwise utility estimation is already fast, so the reduction in total time is smaller (6.4\%).
Using an aggregate-to-fine approach has a lower loss of quality and still reduces the total translation time by some amount.

Reference aggregation is thus a successful strategy to overcome the quadratic complexity of MBR.
However, it is still slower than beam search, as the cost of sampling is now the dominant factor.
Future work could focus on sampling efficiency, e.g., by using fewer hypotheses, improved caching, or speculative sampling approaches~\cite{pmlr-v202-leviathan23a,chen2023accelerating}.

\section{Conclusion}
We proposed reference aggregation, a technique that boosts the efficiency of MBR decoding by shifting the MC sampling from the utility estimation to the reference representation.
Experiments on machine translation showed that reference aggregation speeds up utility estimation by 82\% for \chrf{} and 95–99\% for \comet{} while minimally affecting translation quality.
This reduces the gap to beam search and makes MBR more practical for large-scale applications.

\section*{Limitations}
This work has two main limitations:
\begin{enumerate}
    \item Reference aggregation requires a utility metric based on averageable representations.
    \item For trained metrics, the effectiveness of aggregation needs to be evaluated empirically.
\end{enumerate}
We have demonstrated that reference aggregation is a viable technique for MBR with \chrf{} and \comet{}, leading to a considerable speed-up with minor quality losses.
In the case of \chrf{}, reference aggregation entails a slight modification of the metric definition, but is otherwise exact and not an approximation.
We thus expect that reference aggregation could be applied in a straightforward manner to other lexical overlap metrics such as \textsc{ChrF++}~\cite{popovic-2017-chrf} and \textsc{BLEU}~\cite{papineni-etal-2002-bleu}.

For \comet{}, which is a trained metric, reference aggregation involves the averaging of fixed-size sentence embeddings.
We empirically studied the loss of accuracy incurred by this averaging and found that there is a favorable trade-off between speed and accuracy for the \comet{} models we evaluated.
We recommend that future work validates the effectiveness of reference aggregation for other trained metrics.

While \chrf{} and \comet{} are among the most commonly used metrics for MBR, previous work has also proposed metrics that are not based on averageable reference representations.
For example, \bleurt{}~\cite{sellam-etal-2020-bleurt}, a trained metric that was shown to be effective for MBR~\cite{freitag-etal-2022-high}, is based on a cross-encoder architecture that creates a joint representation for each hypothesis–reference pair.
Future work could investigate in what form, if at all, reference aggregation can be applied to cross-encoder architectures.

Finally, this work studies MBR decoding with a classical sequence-to-sequence NMT model and in the context of sentence-level MT\@.
While MBR decoding has also been successfully applied to MT with large language models~\cite{farinhas-etal-2023-empirical}, more research is needed on MBR decoding with large language models, especially on the document level.

\section*{Acknowledgments}
We thank Clara Meister and Bryan Eikema for helpful discussions and feedback.
This work was funded by the Swiss National Science Foundation (project MUTAMUR; no.~213976).

\bibliography{bibliography}

\clearpage

\appendix

\section{Formal Definition of Reference~Aggregation for ChrF}
\label{appendix:chrf}
The \chrf{} metric~\cite{popovic-2015-chrf} is a harmonic mean of precision and recall scores:
\begin{equation}
    \textsc{ChrF}_\beta = \frac{(1 + \beta^2) \cdot \textsc{ChrP} \cdot \textsc{ChrR}}{\beta^2 \cdot \textsc{ChrP} + \textsc{ChrR}}.
\end{equation}
Internally, \chrf{} converts hypotheses and references into bags of character n-grams.
Such bags can be represented as multisets~(\citealp{knuth1997art}, Section 4.6.3) or as~(sparse) vectors.
We will use vector notation in this formal definition, which allows us to define reference aggregation with standard vector operations.

Let $\textbf{\textit{hyp}} \in \mathbb{R}^{|\mathcal{V}|}$ and $\textbf{\textit{ref}} \in \mathbb{R}^{|\mathcal{V}|}$ be bags representing a hypothesis and a reference, where $\mathcal{V}$ is the vocabulary of all character n-grams up to maximum order~$n$, and the entries $\textbf{\textit{hyp}}_j$ and $\textbf{\textit{ref}}_j$ are the counts of n-gram $j \in \mathcal{V}$ in the hypothesis and reference, respectively.

For a given n-gram order $i \in \{1, \ldots, n\}$, precision and recall are defined as:
\begin{align}
    \label{eq:chrf-precision}
    \textsc{ChrP}_i(\textbf{\textit{hyp}},~\textbf{\textit{ref}}) &= \frac{\sum_{j \in \mathcal{V}_i} \min(\textbf{\textit{hyp}}_j, \textbf{\textit{ref}}_j)}{\sum_{j \in \mathcal{V}_i} \textbf{\textit{hyp}}_j}, \\[1em]
    \label{eq:chrf-recall}
    \textsc{ChrR}_i(\textbf{\textit{hyp}},~\textbf{\textit{ref}}) &= \frac{\sum_{j \in \mathcal{V}_i} \min(\textbf{\textit{hyp}}_j, \textbf{\textit{ref}}_j)}{\sum_{j \in \mathcal{V}_i} \textbf{\textit{ref}}_j},
\end{align}
where $\mathcal{V}_i$ is the set of all character n-grams of order~$i$.
Overall precision and recall are calculated as the arithmetic mean of the precision and recall scores for each n-gram order:
\begin{align}
    \textsc{ChrP}(\textbf{\textit{hyp}},~\textbf{\textit{ref}}) &= \frac{1}{n} \sum_{i=1}^n \textsc{ChrP}_i(\textbf{\textit{hyp}},~\textbf{\textit{ref}}), \\
    \textsc{ChrR}(\textbf{\textit{hyp}},~\textbf{\textit{ref}}) &= \frac{1}{n} \sum_{i=1}^n \textsc{ChrR}_i(\textbf{\textit{hyp}},~\textbf{\textit{ref}}).
\end{align}

When \chrf{} is used as a utility metric in a standard MBR setting, the expected utility of a hypothesis is estimated based on a set $\{\textbf{\textit{ref}}^{(1)}, \ldots, \textbf{\textit{ref}}^{(m)}\}$ of $m$ references:
\begin{equation}
    \text{utility}(\textbf{\textit{hyp}}) = \frac{1}{m} \sum_{k=1}^m \textsc{ChrF}_\beta(\textbf{\textit{hyp}},~\textbf{\textit{ref}}^{(k)}).
\end{equation}

In contrast, reference aggregation first calculates the arithmetic mean of the reference bags:
\begin{equation}
    \overline{\textbf{\textit{ref}}} = [\frac{1}{m} \sum_{k=1}^m \textbf{\textit{ref}}_{1}^{(k)}, \ldots, \frac{1}{m} \sum_{k=1}^m \textbf{\textit{ref}}_{|\mathcal{V}|}^{(k)}],
\end{equation}
and estimates the utility as:
\begin{equation}
    \text{utility}_\text{agg}(\textbf{\textit{hyp}}) = \textsc{ChrF}_\beta(\textbf{\textit{hyp}},~\overline{\textbf{\textit{ref}}}).
\end{equation}

\bigskip
\bigskip

\section{Formal Definition of Partial~Aggregation}
\label{appendix:partial-aggregation}
We conceptualize partial aggregation as follows:
\begin{enumerate}[itemsep=0pt, parsep=0pt]
    \item The set of individual references contains $m$ references.
    \item We randomly partition the set of references into $s$ groups of equal size.
    \item Each group is combined into an average reference representation, resulting in $s$ aggregate references $\overline{\textit{ref}}^{(1)}, \ldots, \overline{\textit{ref}}^{(s)}$.
\end{enumerate}
The expected utility of each sampled hypothesis is then approximated as the average metric score over all aggregate references:
\begin{equation}
    \text{utility}(\textit{hyp}) \approx \frac{1}{s} \sum_{i=1}^s \text{metric}(\textit{hyp}, \overline{\textit{ref}}^{(i)}).
\end{equation}

Like with N-by-S MBR, the parameter $s$ can be seen as the \textit{number of effective references} that determines the computational complexity of the utility estimation.
The case $s=m$ corresponds to standard MBR, where each sampled hypothesis is compared to each reference in a pairwise fashion.
The case $s=1$ corresponds to the full aggregation approach, where a single aggregate reference is created from all references.

\vfill

\clearpage

\section{Formal Definition of Aggregate-to-fine MBR}
\label{appendix:aggregate-to-fine}
Aggregate-to-fine MBR is a special case of coarse-to-fine MBR~\cite{eikema-aziz-2022-sampling}, which uses a cheap proxy utility function to prune the number of hypotheses.
In the case of aggregate-to-fine MBR, the proxy utility function is based on an aggregate reference representation.

\newcommand{\ycf}{\ensuremath{y^{\text{C2F}}}}
\newcommand{\hypset}{\ensuremath{\bar{\mathcal{H}}(x)}}
\newcommand{\hypsetT}{\ensuremath{\bar{\mathcal{H}}_T(x)}}
The general definition of coarse-to-fine MBR is as follows:
Given the original set of sampled hypotheses \(\hypset\) and a proxy utility function \(u_{\text{proxy}}\), coarse-to-fine MBR selects a subset of $T$ hypotheses:
\begin{equation}
    \hypsetT \coloneqq \optop_{\textit{hyp} \in \hypset} ~ u_{\text{proxy}}(\textit{hyp}).
\end{equation}
In the second step, the utility of each hypothesis in the pruned set is estimated using the fine-grained utility function $u_{\text{target}}$:
\begin{equation}
    \ycf \coloneqq \argmax_{\textit{hyp} \in \hypsetT} ~ u_{\text{target}}(\textit{hyp}).
\end{equation}

When experimenting with aggregate-to-fine MBR, we re-use the same utility metric for both steps, but first with an aggregate reference and then with the full set of references:
\begin{align}
    u_{\text{proxy}}(\textit{hyp}) &= \text{metric}(\textit{hyp}, \overline{\textit{ref}}), \\
    u_{\text{target}}(\textit{hyp}) &= \frac{1}{m} \sum_{\textit{ref} \in \textit{refs}} \text{metric}(\textit{hyp}, \textit{ref}).
\end{align}

Note that using the same metric in both steps is not strictly necessary, but has the advantage that the features (e.g., embeddings) only need to be computed once.

\bigskip
\bigskip

\section{Complexity Analysis}
\label{sec:appendix-complexity}

Generally, reference aggregation reduces the complexity of utility estimation from $\bigO(nm)$ to $\bigO(n+m)$, where $n$ is the number of hypotheses and $m$ is the number of references.
The exact complexity depends on the specifics of the utility metric.
Here, we provide a more detailed analysis for \chrf{} and \comet{}.

Above, we stated that utility estimation with these metrics usually has two stages: feature extraction and scoring.
The feature extraction stage is not affected by reference aggregation, and previous work has already remarked that reference features can be extracted once and re-used for all hypotheses~\cite{amrhein-sennrich-2022-identifying}.
If the reference set is identical to the set of hypotheses, the feature extraction stage is in $\bigO(n)$, otherwise $\bigO(n+m)$.

The scoring stage of \chrf{} is dominated by the element-wise minimum function in Eqs.~\ref{eq:chrf-precision} and~\ref{eq:chrf-recall} (or, if the bags of n-grams are represented as multisets, by the intersection operation $\textit{hyp} \cap \textit{ref}$).
Because this operation is performed separately for each hypothesis–reference pair, the complexity is in $\bigO(nm)$.
Reference aggregation reduces the complexity to $\bigO(n+m)$, given that the aggregate reference can be computed once and then re-used for all hypotheses.\footnote{For \chrf{}, reference aggregation can result in an aggregate bag of n-grams that is larger than the bags of the individual references; in the theoretical worst case, where all the references are disjoint, even in an aggregate bag that is $m$ times larger. However, this is a highly unlikely scenario in practice, since different translations of the same source will have substantial overlap, and even if $|\overline{\textit{ref}}| \gg |\textit{ref}|$, the cost of intersection only depends on $|\textit{hyp}|$, assuming that a constant-time hash table is used to check whether each item in $\textit{hyp}$ is contained in $\overline{\textit{ref}}$.}

The same analysis applies to \comet{}.
With standard MBR, Eq.~\ref{eq:comet-score} is evaluated for each hypothesis–reference pair; with reference aggregation, it is only evaluated once for each hypothesis.
The aggregate reference embeddings can be computed once and re-used for all hypotheses.

In practice, the runtime of utility estimation is affected by additional factors.
There may be duplicates among the samples, so the number of scores that effectively need to be computed can vary.
In addition, most aspects of utility estimation can be computed in parallel, which makes the effective runtime highly implementation-dependent.

\clearpage
\onecolumn

\section{Data Statistics}
\label{sec:appendix-data-statistics}
\begin{table*}[htb]
\centering
\begin{tabularx}{\textwidth}{@{}Xrrr@{}}
\toprule
& \# Segments & \# Samples per segment & \# Unique samples per segment \\
\midrule
\textit{newstest21}~ \textsc{en–de} & 1002 & 1024 & 874.2 \\
\textit{newstest21}~ \textsc{de–en} & 1000 & 1024 & 716.9 \\
\textit{newstest21}~ \textsc{en–ru} & 1002 & 1024 & 896.7 \\
\textit{newstest21}~ \textsc{ru–en} & 1000 & 1024 & 727.3 \\
\addlinespace
\textit{newstest22}~ \textsc{en–de} & 2037 & 1024 & 697.5 \\
\textit{newstest22}~ \textsc{de–en} & 1984 & 1024 & 671.4 \\
\textit{newstest22}~ \textsc{en–ru} & 2037 & 1024 & 750.2 \\
\textit{newstest22}~ \textsc{ru–en} & 2016 & 1024 & 726.3 \\
\bottomrule
\end{tabularx}
\caption{Statistics for the datasets used in this paper.
We sample 1024 hypotheses per source segment using epsilon sampling and find that most of the samples are unique.
}
\end{table*}

\vfill

\section{Extended Test Results}
\label{sec:appendix-test-results}
\begin{table*}[htb]
\centering
\begin{tabularx}{\textwidth}{@{}Xrrrrr@{}}
\toprule
& \small{\chrf{}} & \small{\cometinho{}} & \small{\comettt{}} & \small{\cometxl{}} & \small{\bleurtt{}} \\
\midrule
Beam search (size 5) & 58.6 & 56.0 & 84.3 & 92.2 & 73.2 \\
Epsilon sampling ($\epsilon=0.02$) & 52.6 & 45.3 & 81.9 & 89.4 & 70.3 \\
\midrule
MBR with \chrf{} metric & & & & & \\
– standard MBR & 59.8 & 58.3 & 84.5 & 91.8 & 73.4 \\
– reference aggregation & \underline{59.8} & \underline{58.2} & \underline{84.5} & \underline{91.7} & \underline{73.4} \\
– aggregate-to-fine MBR & \underline{59.8} & \underline{58.3} & \underline{84.5} & \underline{91.8} & \underline{73.5} \\
\midrule
MBR with \cometinho{} metric & & & & & \\
– standard MBR & 57.5 & 65.1 & 85.1 & 92.5 & 74.0 \\
– reference aggregation & \textbf{57.8} & 64.5 & 85.0 & 92.4 & 73.9 \\
– aggregate-to-fine MBR & \underline{57.5} & \underline{65.0} & 85.1 & \underline{92.5} & 74.0 \\
\midrule
MBR with \comettt{} metric & & & & & \\
– standard MBR & 57.3 & 60.8 & 87.1 & 93.7 & 74.9 \\
– reference aggregation & \textbf{57.7} & \underline{60.8} & 86.8 & 93.4 & 74.7 \\
– aggregate-to-fine MBR & \textbf{57.4} & \underline{60.8} & 87.0 & \underline{93.7} & 74.9 \\
\midrule
Coarse-to-fine MBR & & & & & \\
– standard \chrf{} to \comettt{} & \textbf{59.3} & 60.1 & 85.8 & 93.0 & 74.4 \\
– aggregate \chrf{} to \comettt{} & \textbf{59.4} & 60.2 & 85.8 & 93.0 & 74.4 \\
\bottomrule
\end{tabularx}
\caption{Extended results on \textit{newstest22} with 1024 samples/references for MBR.
In this table, we include \cometinho{}~\cite{rei-etal-2022-searching} as utility metric, which is a distilled \comet{} model.
Furthermore, as an additional evaluation metric, we report \cometxl{}~\cite{guerreiro2023xcomet}.
We average the evaluation scores across the four translation directions.
\underline{Underline}: no significant difference to standard MBR; \textbf{bold}: significantly better than standard MBR (bootstrap test, $p<0.05$).
}
\label{tab:test-results-extended}
\end{table*}

\vfill

\clearpage

\section{Test Results with 256 Samples}
\label{sec:appendix-test-results-256}
\begin{table*}[htb]
  \centering
\begin{tabularx}{\textwidth}{@{}Xrrrrr@{\hskip 0.3in}r@{}}
\toprule
& \textsc{en--de} & \textsc{de--en} & \textsc{en--ru} & \textsc{ru--en} & Avg. & Time (utility / total) \\
\midrule
Beam search (size 5) & 76.16 & 72.56 & 68.50 & 75.47 & 73.17 &  - ~/ \phantom{0}0.2\,s \\
Epsilon sampling ($\epsilon=0.02$) & 73.39 & 69.70 & 65.79 & 72.13 & 70.25 & - ~/ \phantom{0}0.2\,s \\
\midrule
\mbox{MBR with \chrf{} metric} &  &  &  &  &  &  \\
– standard MBR & 75.90 & 72.66 & 69.27 & 75.60 & 73.36 & \phantom{0}0.1\,s ~/ \phantom{0}1.4\,s \\
– reference aggregation & 75.83 & 72.69 & 69.19 & 75.53 & 73.31 & $<0.1$\,s ~/ \phantom{0}1.3\,s \\
– aggregate-to-fine MBR & 75.90 & 72.67 & 69.29 & 75.58 & 73.36 & $<0.1$\,s ~/ \phantom{0}1.3\,s \\
\midrule
\mbox{MBR with \comettt{} metric} & & & & &  \\
– standard MBR & 77.44 & 73.38 & 72.15 & 76.07 & 74.76 & \phantom{0}1.6\,s ~/ \phantom{0}2.9\,s \\
– reference aggregation & 77.18 & 73.24 & 71.85 & 75.98 & 74.56 & \phantom{0}0.3\,s ~/ \phantom{0}1.6\,s \\
– aggregate-to-fine MBR & 77.42 & 73.36 & 71.98 & 76.05 & 74.70 & \phantom{0}0.4\,s ~/ \phantom{0}1.7\,s \\
\bottomrule
\end{tabularx}
  \caption{Version of Table~\ref{tab:test-results} that uses 256 samples/references for MBR.
  }
  \label{tab:test-results-256-samples}
\end{table*}

\vfill

\section{Effect of Larger Beam Size}

\begin{table*}[htb]
\centering
\begin{tabularx}{0.6\textwidth}{@{}rrrrrr@{}}
\toprule
Beam size & \textsc{en--de} & \textsc{de--en} & \textsc{en--ru} & \textsc{ru--en} & Avg. \\
\midrule
\textbf{5} & 76.16 & 72.56 & 68.50 & 75.47 & 73.17 \\
\textbf{10} & 76.20 & 72.57 & 67.92 & 75.51 & 73.05 \\
\textbf{15} & 76.19 & 72.53 & 68.10 & 75.48 & 73.08 \\
\textbf{20} & 76.18 & 72.54 & 67.84 & 75.49 & 73.01 \\
\textbf{25} & 76.19 & 72.50 & 67.82 & 75.46 & 72.99 \\
\bottomrule
\end{tabularx}
\caption{Increasing the beam size to values larger than 5 does not tend to improve translation quality of beam search on \textit{newstest22} in terms of \bleurtt{}.
}
\end{table*}

\vfill
\vfill
\vfill

\clearpage

\section{Top-1 Accuracy of Efficiency Methods}
\label{sec:appendix-top1-accuracy}

\begin{figure}[H]
\hspace*{0.1cm}
\textbf{Utility metric: \chrf{}}
\vspace*{-0.1cm}
\begin{tikzpicture}
\pgfplotsset{width=0.1975\textwidth}
\begin{groupplot}[
    group style={
        group size=4 by 1,
        xlabels at=edge bottom,
        ylabels at=edge left,
        horizontal sep=0.5cm,
    },
    scale only axis,
    xmode=log,
    log basis x={2},
    xlabel={Number of effective references (log scale)},
    xlabel style={xshift=1.6cm, yshift=-0.2cm},
    xmin=1, xmax=1024,
    xtick={1,2,4,8,16,32,64,128,256,512,1024},
    xticklabels={\scriptsize{$\dfrac{n}{1}$}, \scriptsize{$\dfrac{n}{2}$}, \scriptsize{$\dfrac{n}{4}$}, \scriptsize{$\dfrac{n}{8}$},,,,\vphantom{\scriptsize{$\dfrac{n}{}$}}\ldots,,,\vphantom{\small{$\dfrac{n}{n}$}}\small{$1$}},
    ymin=0, ymax=1,
    ytick={0, 0.5, 1},
    yticklabels={0\%, 50\%, 100\%},
    grid style=dashed,
    height=2cm
]

% Plot 1 (en-de accuracy)
\nextgroupplot[
    title=\textsc{en--de},
    title style={yshift=-5pt},
    ylabel={Top-1 accuracy},
    ylabel near ticks,
    ylabel style={align=center},
    ytick pos=left,
    legend style={at={(3.52,-0.674)},anchor=south, draw=none, fill=none, style={/tikz/every even column/.append style={column sep=0.2cm}}},
    legend columns=-1
]
\addplot[color=uzhblue, mark=*] coordinates {
(1,1.00000)(2,0.93613)(4,0.90419)(8,0.87625)(16,0.85529)(32,0.84331)(64,0.84032)(128,0.83932)(256,0.83533)(512,0.83533)(1024,0.83234)
};
\addlegendentry{Reference aggregation}

\addplot[color=uzhorange, mark=o] coordinates {
(1,1.00000)(2,0.85828)(4,0.76647)(8,0.64371)(16,0.54890)(32,0.43912)(64,0.26846)(128,0.17166)(256,0.06687)(512,0.02894)(1024,0.01397)
};
\addlegendentry{N-by-S}

% Plot 2 (de-en accuracy)
\nextgroupplot[title=\textsc{de--en}, title style={yshift=-5pt}, yticklabels={}, ytick pos=left, xlabel=\empty]
\addplot[color=uzhblue, mark=*] coordinates {
(1,1.00000)(2,0.92400)(4,0.88500)(8,0.86100)(16,0.85000)(32,0.85000)(64,0.84700)(128,0.84600)(256,0.84200)(512,0.84100)(1024,0.84300)
};
\addplot[color=uzhorange, mark=o] coordinates {
(1,1.00000)(2,0.84900)(4,0.77300)(8,0.68600)(16,0.56600)(32,0.46000)(64,0.32300)(128,0.19300)(256,0.10800)(512,0.05000)(1024,0.05100)
};

% Plot 3 (en-ru accuracy)
\nextgroupplot[title=\textsc{en--ru}, title style={yshift=-5pt}, yticklabels={}, ytick pos=left, xlabel=\empty]
\addplot[color=uzhblue, mark=*] coordinates {
(1,1.00000)(2,0.91018)(4,0.86028)(8,0.84331)(16,0.81637)(32,0.81038)(64,0.80339)(128,0.79741)(256,0.79641)(512,0.79441)(1024,0.79441)
};
\addplot[color=uzhorange, mark=o] coordinates {
(1,1.00000)(2,0.85329)(4,0.76347)(8,0.65669)(16,0.52894)(32,0.38922)(64,0.23353)(128,0.13573)(256,0.05190)(512,0.02196)(1024,0.01497)
};

% Plot 4 (ru-en accuracy)
\nextgroupplot[title=\textsc{ru--en}, title style={yshift=-5pt}, yticklabels={}, ytick pos=left, xlabel=\empty]
\addplot[color=uzhblue, mark=*] coordinates {
(1,1.00000)(2,0.94500)(4,0.90900)(8,0.87900)(16,0.85600)(32,0.85000)(64,0.85100)(128,0.84800)(256,0.84600)(512,0.84600)(1024,0.84500)
};
\addplot[color=uzhorange, mark=o] coordinates {
(1,1.00000)(2,0.86300)(4,0.79700)(8,0.69600)(16,0.58400)(32,0.48400)(64,0.31500)(128,0.21500)(256,0.11100)(512,0.04800)(1024,0.04000)
};
\end{groupplot}
\end{tikzpicture}
\vspace*{-0.5cm}
\hspace*{0.1cm}
\textbf{Utility metric: \comettt{}}
\vspace*{-0.1cm}
\begin{tikzpicture}
\pgfplotsset{width=0.1975\textwidth}
\begin{groupplot}[
    group style={
        group size=4 by 1,
        xlabels at=edge bottom,
        ylabels at=edge left,
        horizontal sep=0.5cm,
    },
    scale only axis,
    xmode=log,
    log basis x={2},
    xlabel={Number of eff. references (log scale)},
    xlabel style={xshift=1.21cm, yshift=-0.2cm},
    xmin=1, xmax=1024,
    xtick={1,2,4,8,16,32,64,128,256,512,1024},
    xticklabels={\scriptsize{$\dfrac{n}{1}$}, \scriptsize{$\dfrac{n}{2}$}, \scriptsize{$\dfrac{n}{4}$}, \scriptsize{$\dfrac{n}{8}$},,,,\vphantom{\scriptsize{$\dfrac{n}{}$}}\ldots,,,\vphantom{\small{$\dfrac{n}{n}$}}\small{$1$}},
    ymin=0, ymax=1,
    ytick={0, 0.5, 1},
    yticklabels={0\%, 50\%, 100\%},
    grid style=dashed,
    height=2cm
]

% Plot 1 (en-de accuracy)
\nextgroupplot[
    title=\textsc{en--de},
    title style={yshift=-5pt},
    ylabel={Top-1 accuracy},
    ylabel near ticks,
    ylabel style={align=center},
    ytick pos=left,
    legend style={at={(3.22,-0.674)},anchor=south, draw=none, fill=none, style={/tikz/every even column/.append style={column sep=0.2cm}}},
    legend columns=-1
]
\addplot[color=uzhblue, mark=*] coordinates {
(1,1.00000)(2,0.79741)(4,0.67265)(8,0.60978)(16,0.55988)(32,0.53992)(64,0.52595)(128,0.51697)(256,0.51397)(512,0.51098)(1024,0.51397)
};
\addlegendentry{Reference aggregation}

\addplot[color=uzhorange, mark=o] coordinates {
(1,1.00000)(2,0.92814)(4,0.87525)(8,0.81038)(16,0.73653)(32,0.65269)(64,0.51497)(128,0.36527)(256,0.18563)(512,0.04391)(1024,0.01098)
};
\addlegendentry{N-by-S}

\addplot[dashed, gray] coordinates {
(1,0.05888)(2,0.05788)(4,0.05289)(8,0.05289)(16,0.05389)(32,0.05689)(64,0.05589)(128,0.05589)(256,0.05589)(512,0.05589)(1024,0.05689)
};
\addlegendentry{\chrf{}}

% Plot 2 (de-en accuracy)
\nextgroupplot[title=\textsc{de--en}, title style={yshift=-5pt}, yticklabels={}, ytick pos=left, xlabel=\empty]
\addplot[color=uzhblue, mark=*] coordinates {
(1,1.00000)(2,0.83800)(4,0.72000)(8,0.65800)(16,0.62200)(32,0.60500)(64,0.59200)(128,0.59300)(256,0.58800)(512,0.59000)(1024,0.58700)
};
\addplot[color=uzhorange, mark=o] coordinates {
(1,1.00000)(2,0.92500)(4,0.84900)(8,0.77800)(16,0.71700)(32,0.64100)(64,0.50000)(128,0.35300)(256,0.19600)(512,0.07300)(1024,0.04200)
};
\addplot[dashed, gray] coordinates {
(1,0.13900)(2,0.13700)(4,0.13200)(8,0.13300)(16,0.13300)(32,0.13200)(64,0.13300)(128,0.13300)(256,0.13300)(512,0.13400)(1024,0.13400)
};

% Plot 3 (en-ru accuracy)
\nextgroupplot[title=\textsc{en--ru}, title style={yshift=-5pt}, yticklabels={}, ytick pos=left, xlabel=\empty]
\addplot[color=uzhblue, mark=*] coordinates {
(1,1.00000)(2,0.80439)(4,0.69361)(8,0.61677)(16,0.56587)(32,0.54491)(64,0.53393)(128,0.52595)(256,0.52196)(512,0.52096)(1024,0.52395)
};
\addplot[color=uzhorange, mark=o] coordinates {
(1,1.00000)(2,0.91916)(4,0.88323)(8,0.82335)(16,0.73852)(32,0.62475)(64,0.49900)(128,0.38423)(256,0.22754)(512,0.06487)(1024,0.01098)
};
\addplot[dashed, gray] coordinates {
(1,0.04491)(2,0.04391)(4,0.04391)(8,0.04391)(16,0.04391)(32,0.04391)(64,0.04391)(128,0.04391)(256,0.04391)(512,0.04391)(1024,0.04391)
};

% Plot 4 (ru-en accuracy)
\nextgroupplot[title=\textsc{ru--en}, title style={yshift=-5pt}, yticklabels={}, ytick pos=left, xlabel=\empty]
\addplot[color=uzhblue, mark=*] coordinates {
(1,1.00000)(2,0.83800)(4,0.72200)(8,0.65000)(16,0.62100)(32,0.60300)(64,0.59200)(128,0.58400)(256,0.58300)(512,0.58400)(1024,0.58600)
};
\addplot[color=uzhorange, mark=o] coordinates {
(1,1.00000)(2,0.91300)(4,0.85800)(8,0.79900)(16,0.70200)(32,0.62200)(64,0.46600)(128,0.31500)(256,0.16900)(512,0.05200)(1024,0.03800)
};
\addplot[dashed, gray] coordinates {
(1,0.12500)(2,0.12200)(4,0.12000)(8,0.12100)(16,0.12300)(32,0.12400)(64,0.12400)(128,0.12400)(256,0.12400)(512,0.12400)(1024,0.12400)
};

\end{groupplot}
\end{tikzpicture}
\vspace*{-0.5cm}
\caption{Top-1 accuracy of MBR efficiency methods on \textit{newstest21}, analogous to Figure~\ref{fig:validation-top20-accuracy}.}
\label{fig:validation-n-by-s-accuracy-top1}
\end{figure}

\vfill

\section{Validation Results for Cometinho}

\begin{figure}[H]
\hspace*{0.1cm}
\textbf{Top-20 accuracy}
\vspace*{-0.1cm}
\begin{tikzpicture}
\pgfplotsset{width=0.1975\textwidth}
\begin{groupplot}[
    group style={
        group size=4 by 1,
        xlabels at=edge bottom,
        ylabels at=edge left,
        horizontal sep=0.5cm,
    },
    scale only axis,
    xmode=log,
    log basis x={2},
    xlabel={Number of eff. references (log scale)},
    xlabel style={xshift=1.21cm, yshift=-0.2cm},
    xmin=1, xmax=1024,
    xtick={1,2,4,8,16,32,64,128,256,512,1024},
    xticklabels={\scriptsize{$\dfrac{n}{1}$}, \scriptsize{$\dfrac{n}{2}$}, \scriptsize{$\dfrac{n}{4}$}, \scriptsize{$\dfrac{n}{8}$},,,,\vphantom{\scriptsize{$\dfrac{n}{}$}}\ldots,,,\vphantom{\small{$\dfrac{n}{n}$}}\small{$1$}},
    ymin=0, ymax=1,
    ytick={0, 0.5, 1},
    yticklabels={0\%, 50\%, 100\%},
    grid style=dashed,
    height=1.5cm
]

% Plot 1 (en-de accuracy)
\nextgroupplot[
    title=\textsc{en--de},
    title style={yshift=-5pt},
    ylabel={Top-20 accuracy},
    ylabel near ticks,
    ylabel style={align=center},
    ytick pos=left,
    legend style={at={(3.22,-0.874)},anchor=south, draw=none, fill=none, style={/tikz/every even column/.append style={column sep=0.2cm}}},
    legend columns=-1
]
\addplot[color=uzhblue, mark=*] coordinates {
(1,1.00000)(2,0.98703)(4,0.97804)(8,0.97405)(16,0.96507)(32,0.96208)(64,0.96108)(128,0.95908)(256,0.95808)(512,0.95808)(1024,0.95808)
};
\addlegendentry{Reference aggregation}

\addplot[color=uzhorange, mark=o] coordinates {
(1,1.00000)(2,0.98902)(4,0.98104)(8,0.97605)(16,0.95110)(32,0.89721)(64,0.83433)(128,0.69661)(256,0.49301)(512,0.27345)(1024,0.11377)
};
\addlegendentry{N-by-S}

\addplot[dashed, gray] coordinates {
(1,0.26946)(2,0.26647)(4,0.26447)(8,0.26447)(16,0.26547)(32,0.26647)(64,0.26747)(128,0.26447)(256,0.26547)(512,0.26547)(1024,0.26447)
};
\addlegendentry{\chrf{}}

% Plot 2 (de-en accuracy)
\nextgroupplot[title=\textsc{de--en}, title style={yshift=-5pt}, yticklabels={}, ytick pos=left, xlabel=\empty]
\addplot[color=uzhblue, mark=*] coordinates {
(1,1.00000)(2,0.97900)(4,0.96100)(8,0.94900)(16,0.94700)(32,0.94200)(64,0.94100)(128,0.94100)(256,0.94200)(512,0.94100)(1024,0.94100)
};
\addplot[color=uzhorange, mark=o] coordinates {
(1,1.00000)(2,0.98400)(4,0.95700)(8,0.91900)(16,0.89200)(32,0.83500)(64,0.75100)(128,0.58800)(256,0.39300)(512,0.22100)(1024,0.10200)
};
\addplot[dashed, gray] coordinates {
(1,0.34100)(2,0.34200)(4,0.33800)(8,0.33900)(16,0.34000)(32,0.34000)(64,0.33900)(128,0.33900)(256,0.33900)(512,0.34100)(1024,0.34200)
};

% Plot 3 (en-ru accuracy)
\nextgroupplot[title=\textsc{en--ru}, title style={yshift=-5pt}, yticklabels={}, ytick pos=left, xlabel=\empty]
\addplot[color=uzhblue, mark=*] coordinates {
(1,1.00000)(2,0.98503)(4,0.98204)(8,0.97804)(16,0.97305)(32,0.96906)(64,0.96806)(128,0.96806)(256,0.96707)(512,0.96707)(1024,0.96707)
};
\addplot[color=uzhorange, mark=o] coordinates {
(1,1.00000)(2,0.99301)(4,0.98802)(8,0.97705)(16,0.96008)(32,0.92415)(64,0.86627)(128,0.75349)(256,0.52994)(512,0.32834)(1024,0.12874)
};
\addplot[dashed, gray] coordinates {
(1,0.24251)(2,0.24052)(4,0.24351)(8,0.24251)(16,0.24052)(32,0.24052)(64,0.24052)(128,0.24152)(256,0.24152)(512,0.24152)(1024,0.24152)
};

% Plot 4 (ru-en accuracy)
\nextgroupplot[title=\textsc{ru--en}, title style={yshift=-5pt}, yticklabels={}, ytick pos=left, xlabel=\empty]
\addplot[color=uzhblue, mark=*] coordinates {
(1,1.00000)(2,0.98100)(4,0.97200)(8,0.96200)(16,0.95600)(32,0.95100)(64,0.94600)(128,0.94300)(256,0.94200)(512,0.94100)(1024,0.94100)
};
\addplot[color=uzhorange, mark=o] coordinates {
(1,1.00000)(2,0.98000)(4,0.97600)(8,0.93900)(16,0.90900)(32,0.84100)(64,0.73100)(128,0.59300)(256,0.38700)(512,0.20700)(1024,0.09900)
};
\addplot[dashed, gray] coordinates {
(1,0.34200)(2,0.33800)(4,0.33300)(8,0.33700)(16,0.33700)(32,0.33800)(64,0.33700)(128,0.33700)(256,0.33600)(512,0.33600)(1024,0.33600)
};

\end{groupplot}
\end{tikzpicture}
\vspace*{-0.5cm}
\hspace*{0.1cm}
\textbf{Top-1 accuracy}
\vspace*{-0.1cm}
\begin{tikzpicture}
\pgfplotsset{width=0.1975\textwidth}
\begin{groupplot}[
    group style={
        group size=4 by 1,
        xlabels at=edge bottom,
        ylabels at=edge left,
        horizontal sep=0.5cm,
    },
    scale only axis,
    xmode=log,
    log basis x={2},
    xlabel={Number of eff. references (log scale)},
    xlabel style={xshift=1.21cm, yshift=-0.2cm},
    xmin=1, xmax=1024,
    xtick={1,2,4,8,16,32,64,128,256,512,1024},
    xticklabels={\scriptsize{$\dfrac{n}{1}$}, \scriptsize{$\dfrac{n}{2}$}, \scriptsize{$\dfrac{n}{4}$}, \scriptsize{$\dfrac{n}{8}$},,,,\vphantom{\scriptsize{$\dfrac{n}{}$}}\ldots,,,\vphantom{\small{$\dfrac{n}{n}$}}\small{$1$}},
    ymin=0, ymax=1,
    ytick={0, 0.5, 1},
    yticklabels={0\%, 50\%, 100\%},
    grid style=dashed,
    height=2cm
]

% Plot 1 (en-de accuracy)
\nextgroupplot[
    title=\textsc{en--de},
    title style={yshift=-5pt},
    ylabel={Top-1 accuracy},
    ylabel near ticks,
    ylabel style={align=center},
    ytick pos=left,
    legend style={at={(3.22,-0.674)},anchor=south, draw=none, fill=none, style={/tikz/every even column/.append style={column sep=0.2cm}}},
    legend columns=-1
]
\addplot[color=uzhblue, mark=*] coordinates {
(1,1.00000)(2,0.82535)(4,0.72156)(8,0.65569)(16,0.61876)(32,0.58483)(64,0.57485)(128,0.56886)(256,0.55988)(512,0.55788)(1024,0.55888)
};
\addlegendentry{Reference aggregation}

\addplot[color=uzhorange, mark=o] coordinates {
(1,1.00000)(2,0.87924)(4,0.80639)(8,0.71657)(16,0.62275)(32,0.48104)(64,0.34232)(128,0.20459)(256,0.04990)(512,0.01098)(1024,0.00898)
};
\addlegendentry{N-by-S}

\addplot[dashed, gray] coordinates {
(1,0.08483)(2,0.08583)(4,0.08583)(8,0.08483)(16,0.08483)(32,0.08583)(64,0.08483)(128,0.08483)(256,0.08483)(512,0.08483)(1024,0.08483)
};
\addlegendentry{\chrf{}}

% Plot 2 (de-en accuracy)
\nextgroupplot[title=\textsc{de--en}, title style={yshift=-5pt}, yticklabels={}, ytick pos=left, xlabel=\empty]
\addplot[color=uzhblue, mark=*] coordinates {
(1,1.00000)(2,0.85400)(4,0.78200)(8,0.72700)(16,0.69500)(32,0.67000)(64,0.66200)(128,0.65600)(256,0.65600)(512,0.65700)(1024,0.65700)
};
\addplot[color=uzhorange, mark=o] coordinates {
(1,1.00000)(2,0.90600)(4,0.82400)(8,0.74700)(16,0.66900)(32,0.51600)(64,0.39500)(128,0.24400)(256,0.10400)(512,0.05200)(1024,0.04600)
};
\addplot[dashed, gray] coordinates {
(1,0.18600)(2,0.18300)(4,0.18200)(8,0.18400)(16,0.18400)(32,0.18400)(64,0.18300)(128,0.18300)(256,0.18400)(512,0.18500)(1024,0.18500)
};

% Plot 3 (en-ru accuracy)
\nextgroupplot[title=\textsc{en--ru}, title style={yshift=-5pt}, yticklabels={}, ytick pos=left, xlabel=\empty]
\addplot[color=uzhblue, mark=*] coordinates {
(1,1.00000)(2,0.81936)(4,0.71257)(8,0.64471)(16,0.61477)(32,0.58982)(64,0.57884)(128,0.57186)(256,0.57086)(512,0.56886)(1024,0.56487)
};
\addplot[color=uzhorange, mark=o] coordinates {
(1,1.00000)(2,0.89222)(4,0.84431)(8,0.74551)(16,0.67066)(32,0.55190)(64,0.39421)(128,0.24950)(256,0.08483)(512,0.01597)(1024,0.01297)
};
\addplot[dashed, gray] coordinates {
(1,0.07685)(2,0.07585)(4,0.07285)(8,0.07186)(16,0.06786)(32,0.06786)(64,0.06886)(128,0.06986)(256,0.06986)(512,0.06986)(1024,0.06986)
};

% Plot 4 (ru-en accuracy)
\nextgroupplot[title=\textsc{ru--en}, title style={yshift=-5pt}, yticklabels={}, ytick pos=left, xlabel=\empty]
\addplot[color=uzhblue, mark=*] coordinates {
(1,1.00000)(2,0.84200)(4,0.77300)(8,0.71600)(16,0.68400)(32,0.66700)(64,0.65400)(128,0.64800)(256,0.64300)(512,0.63900)(1024,0.63600)
};
\addplot[color=uzhorange, mark=o] coordinates {
(1,1.00000)(2,0.90600)(4,0.82900)(8,0.75000)(16,0.63400)(32,0.52100)(64,0.37500)(128,0.23000)(256,0.10600)(512,0.04300)(1024,0.03700)
};
\addplot[dashed, gray] coordinates {
(1,0.19000)(2,0.18500)(4,0.17900)(8,0.17800)(16,0.17900)(32,0.17800)(64,0.17800)(128,0.17700)(256,0.17800)(512,0.17800)(1024,0.17900)
};

\end{groupplot}
\end{tikzpicture}
\vspace*{-0.5cm}
\caption{Accuracy of MBR efficiency methods on \textit{newstest21} when using the \cometinho{} model~\cite{rei-etal-2022-searching} as utility metric.
}
\label{fig:validation-cometinho}
\end{figure}

\vfill

\end{document}